# Pixel-Anchor: A Fast Oriented Scene Text Detector with Combined Networks

Yuan Li, Yuanjie Yu, Zefeng Li, Yangkun Lin, Meifang Xu, Jiwei Li, Xi Zhou

Cloudwalk Technology, Shanghai, China.
{liyuan, yuyuanjie, lizefeng, linyangkun, xumeifang, lijiwei, zhouxi}@cloudwalk.cn

## Abstract

*Recently, semantic segmentation and general object detection frameworks have been widely adopted by scene text detecting tasks. However, both of them alone have obvious shortcomings in practice. In this paper, we propose a novel end-to-end trainable deep neural network framework, named Pixel-Anchor, which combines semantic segmentation and SSD in one network by feature sharing and anchor-level attention mechanism to detect oriented scene text. To deal with scene text which has large variances in size and aspect ratio, we combine FPN and ASPP operation as our encoder-decoder structure in the semantic segmentation part, and propose a novel Adaptive Predictor Layer in the SSD. Pixel-Anchor detects scene text in a single network forward pass, no complex post-processing other than an efficient fusion Non-Maximum Suppression is involved. We have benchmarked the proposed Pixel-Anchor on the public datasets. Pixel-Anchor outperforms the competing methods in terms of text localization accuracy and run speed, more specifically, on the ICDAR 2015 dataset, the proposed algorithm achieves an F-score of 0.8768 at 10 FPS for 960×1728 resolution images.*

## 1. Introduction

Recently, with the rise of the advances in deep Convolutional Neural Networks (CNNs) [1, 2, 3, 4], detecting text in the natural scene has attracted increasing attention, due to its numerous applications in Ads filtering, scene understanding, product identification, and target geolocation. However, because of large variations in size, orientations, and aspect ratios, as well as complex image degradation, extreme illumination and occlusion, scene text detection is still facing considerable challenges.

As semantic segmentation [5, 6, 7] and general object detection [8, 9, 10] technologies are adopted to scene text detection task, more and more outstanding scene text detection frameworks are emerging. Generally, most of the state-of-the-art methods can be divided into two categories, the pixel-based method [17, 18, 19] from semantic segmentation, and the anchor-based method [13, 14, 22, 23] from general object detection. Different text elements are detected by the two methods, pixel-level ones for the former, and the anchor-level ones for the latter, which are combined with corresponding bounding box regression methods to get final text regions, as shown in Figure 1. Unfortunately, neither of them meets our satisfaction when being simply applied. The pixel-based method has high precision, but it has low recall due to too sparse pixel-level features for small texts. The anchor-based method has high recall because anchor-level features are less sensitive to the text size, but it suffers from "Anchor Matching Dilemma" problem (described in Section 2.2), and it is not so good to get high precision as the pixel-based method does. Moreover, the existing methods perform poorly in detecting long Chinese texts running across the images.

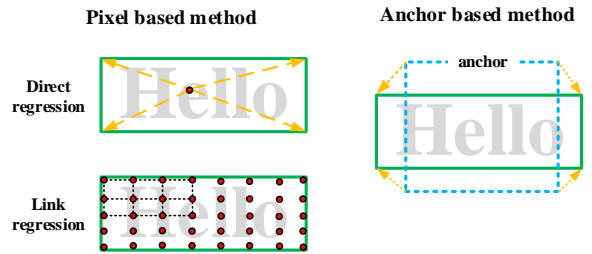

Figure 1: The pixel-based method and the anchor-based method for text bounding box regression.

To address these problems, in this paper, we propose a novel end-to-end trainable deep neural network framework, named Pixel-Anchor, which combines the advantages of the pixel-based and the anchor-based method. The whole framework has two parts, named the pixel-based module and the anchor-based module. The two parts share the features extracted from ResNet-50 [3]. In Pixel-Anchor, the segmentation heat map from the pixel-based module is fed into the anchor-based module by the anchor-level attention mechanism [11], which improves the anchor-based-method accuracy. At the stage of inference, we conduct a fusion Non-Maximum Suppression (NMS) [14] to get the final detections. Specifically, we only keep the small and long anchors in the anchor-based module by anchor trimming and meanwhile remove the small detection boxes in the pixel-based module. Finally, we gather all the remaining detection boxes and conduct a cascaded NMS to get the final detection results. The efficiency of Pixel-Anchor gains resulting from feature sharing between the pixel-based module and the anchor-based module. The whole network can be trained end-to-end by back propagation and stochastic gradient descent.

In the pixel-based module, we combine Feature Pyramid Network (FPN) [11] and atrous spacial pyramid pooling (ASPP) [6, 7] operation as our encoder-decoder structure for semantic segmentation. It is a low-cost way to get large receptive field by conducting ASPP operation on the 1/16 feature map. In the anchor-based module, we adopt SSD [10] as base framework and propose an Adaptive Predictor Layer (APL) to better detect scene text with large variances in size and aspect ratio. The APL is efficient to adjust the receptive field of the network to accommodate the shape of text. To detect long text lines running across the images, we further propose "long anchors" and "anchor density" in the APL.

To show the effectiveness of our proposed Pixel-Anchor framework, we conduct extensive experiments on two competitive benchmark datasets including ICDAR 2015 [15] and ICDAR 2017 MLT [16], Pixel-Anchor prevails against the other methods in terms of text localization accuracy and run speed.

The main contributions of this paper are as follows:

● We introduce Pixel-Anchor, a single-shot oriented scene text detector which combines the advantages of the pixel-based and the anchor-based method by feature sharing, an anchor-level attention mechanism, and a fusion NMS.

● We propose APL for SSD to better detect objects like scene text having large variances in size and aspect ratio. Especially, the APL can effectively detect long text lines running across the images.

● These features lead to simple end-to-end training and improve the speed vs accuracy trade-off. Especially on low resolution input images, our method can still give good performance.

2. Related Works

2.1. The pixel-based method

The stroke characteristics of texts are obvious, so it is easy to segment text pixels from background. The pixel-based method predicts text bounding boxes directly from text pixels.

At the text pixels segmenting stage, a text/non-text score at each pixel is predicted by a typical encoder-decoder network, which is widely used in semantic segmentation tasks. The popular pixel-based methods, EAST [17], Pixel-Link [18], and PSENet [19] use FPN [11] as the encoder-decoder module. In FPN, a U-shape structure is constructed to maintain both high spatial resolution and semantic information.

At text bounding box predicting stage, EAST predicts a text bounding box at each positive text pixel and then conducts the locality-aware NMS to get the final detection results. With a few modifications on EAST, FOTS [20] achieves very competitive results on the public datasets. EAST and FOTS are efficient and accurate, however, because the maximal size of text instances handled by them is proportional to the receptive field of the network, EAST and FOTS perform poorly in detecting very long texts like those running across the images. Pixel-Link and PSENet obtain text instances by linking the adjacent text pixels together. The pixel linking method overcomes the receptive field limitation problem, so it can detect very long text lines. However, the pixel linking method requires a complex post-processing, which is vulnerable to background disturbance.

To improve efficiency, the semantic segmentation methods usually predict the foreground/background score map on the 1/4 or even smaller feature maps. To distinguish very close text instances, the pixel-based methods usually use "shrunk polygons" to assign ground truth text labels, as shown in Figure 2. By assigning text labels to "shrunk polygons", the close text instances A and B can be easily distinguished. Note that PSENet uses progressive scales shrunk polygons to generate ground truth text labels. Since the ground truth labels have been shrunk so much, as for the pixel-based method, the input image has to maintain high resolution to detect small texts and distinguish close small texts at a correspondingly high cost in time.

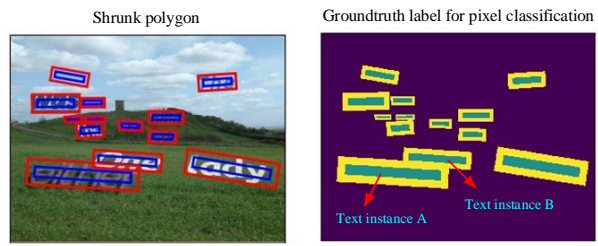

Figure 2: The shrunk polygon label assignments.

2.2. The anchor-based method

The concept of anchor originally comes from Faster-RCNN [8], which is a two-step general object detection framework. First, it generates proposals using anchors (prior default boxes). Secondly, Deep Features describing the proposals are extracted from feature maps via ROI pooling, and object offsets relative to the anchors are regressed based on Deep Features instead of directly predicting the bounding boxes. The anchor-based bounding boxes regression is applied in many popular general object detection frameworks, such as YOLO [9, 21] and SSD [10]. SSD generates a hierarchy of feature pyramids, and places anchors with different scales on each feature map. Among all the anchor-based methods, SSD has a good trade-off between speed and precision and is widely adopted for scene text detection.

Textboxes [13] and Textboxes++ [14] modify SSD to detect natural scene text. Textboxes proposes three special designs for adapting an SSD network to efficiently detect horizontal scene text. One, it adds large aspect ratios anchors to better fit the shape of text boxes. Specifically, Textboxes sets the aspect ratios of anchors to 1, 2, 3, 5, 7, and 10. Two, it increases anchors density by offsetting anchors in the vertical direction to detect dense texts. Three, it adopts 1×5 convolutional filters instead of the standard 3×3 ones. 1×5 convolutional filters yield rectangular receptive field, which better fit words with larger aspect ratios. Based on Textboxes, Textboxes++ enlarges the aspect ratios of default boxes to 1, 2, 3, 5, 1/2, 1/3, 1/5 and uses 3×5 convolutional filters for oriented text detection.

Textboxes++ has achieved good results on the public datasets, but it fails to deal with dense and large-angle texts. Textboxes++ uses horizontal rectangles as anchors, as for two close large-angle text instances, it is hard to determine which text instance should be matched with the anchor. We call the phenomenon "Anchor Matching Dilemma", as shown in Figure 3. This phenomenon prevents the network from performing well in detecting dense large-angle texts.

To overcome the above problem, DMPNet [22] and RRPN [23] use quadrilateral anchors with different orientations to detect oriented texts. However, it greatly increases the number of anchors. Among the methods, it is time-consuming to compute the intersection area between two arbitrary quadrilaterals, especially when the number of anchors is large (one hundred thousand orders of magnitude).

In addition to "Anchor Matching Dilemma", Textboxes++ also suffers from receptive field limitation. The receptive field of 1×5 or 3×5 convolutional filter is not enough to detect text lines running across the images, even the aspect ratio of anchors is enlarged to 1:10, it cannot match the long Chinese text lines whose aspect ratio usually exceeds 1:30.

Compared with the pixel-based method, the anchor-based method directly learns anchor-level abstract features to describe text instances, not pixel-level stroke features as the pixel-based method does. The anchor-level abstract features have to face more diversity, as a result, it usually has more false positives. But also from the anchor-level abstract feature, the anchor-based method is more robust to text size, and is efficient to detect small texts. According to our experiments, when using small images, the anchor-based method usually has higher recall scores than the pixel-based method.

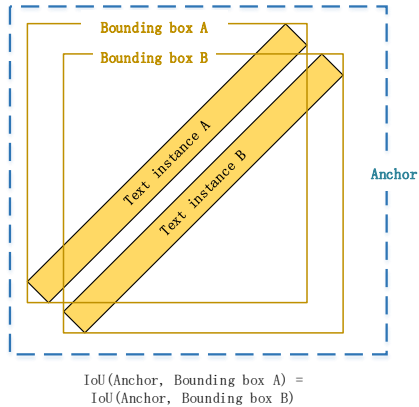

Figure 3: The Anchor Matching Dilemma.

3. Pixel-Anchor

We describe our Pixel-Anchor framework for scene text detection in this Section.

3.1. Overall Architecture

An overview of our framework is illustrated in Figure 4. We combine the pixel-based method and the anchor-based method in one network by feature sharing and anchor-level attention. ResNet-50 [3] is used as the feature extractor backbone. The output stride of ResNet-50 before classification is 32. For the task of semantic segmentation in the pixel-based module, the output stride is set to 16 for denser feature extraction by removing the striding in the last res-block and applying the atrous convolution (rate=2) correspondingly. 1/4, 1/8 and 1/16 feature maps are extracted from the ResNet-50 backbone and shared in both the pixel-based module and the anchor-based module. The segmentation heat map in the pixel-based module is fed to the anchor-based module according to the anchor-level attention mechanism. At the stage of inference, no complex post-processing other than an efficient fusion NMS is involved.

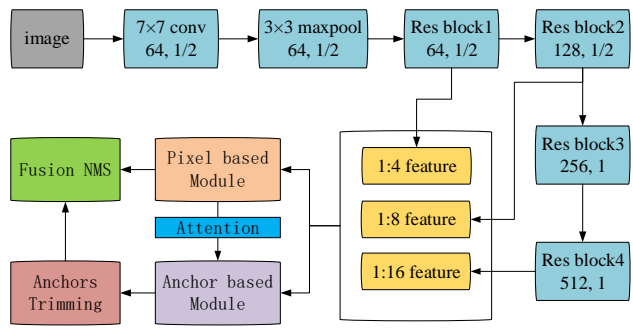

Figure 4: The overall architecture of Pixel-Anchor.

## 3.2. The pixel-based module

Most pixel-based text detectors like EAST and Pixel-Link use FPN as the encoder-decoder module. To increase the receptive field of our network, we combine FPN and ASPP operation as our encoder-decoder structure, and modify ASPP with dilation rate of {3, 6, 9, 12, 15, 18} instead of {6, 12, 18} in DeepLabv3+ [7] for obtaining a finer receptive field. At the decoding stage, the encoder features are first bilinearly upsampled by a factor of 2 and then concatenated with the corresponding low-level features from the network backbone. The decoded feature map maintains both the high spatial resolution and semantic information. The overview of the pixel-based module is illustrated in Figure 5.

Compared with the FPN module, ASPP operation is simpler and more efficient. It is a low-cost way to increase the receptive field of the network, because most of its operations are conducted on the 1/16 feature map. So the network possesses a larger receptive field while maintaining high efficiency.

The output of the pixel-based module consists of two parts: the rotated box and the attention heat map. The rotated box (RBox) predictor contains 6 channels like EAST. The first channel computes the probability of each pixel being a positive text, the following 4 channels predict its distances to top, bottom, left, right of the text bounding box that contains this pixel, and the last channel predicts the orientation of the text bounding box. The attention heat map contains one channel indicating the probability of each pixel being a positive text, and will be fed into the anchor-based module.

For RBox predictor, in order to distinguish very close text instances, the "shrunk polygon" method is used to assign ground truth text labels like FOTS. Only shrunk part of the original text region is considered as the positive text area, while the area between the bounding box and shrunk polygon is ignored. For attention heat map, the "shrunk polygon" method is not applied and all the original text regions are considered as the positive text areas.

We adopt online hard example mining (OHEM) [43] for calculating pixel classification loss. For each image, 512 hard negative non-text pixels, 512 random negative non-text pixels, and all positive text pixels are selected for classification training. Respectively denoting pixel sets selected for classification training in RBox predictor and attention heat map as $\Omega_{RBox}$ and $\Omega_{heatmap}$, the loss function for pixel classification can be formulated as:

$$L_{p\_cls} = \frac{1}{|\Omega_{RBox}|} \sum_{i \in \Omega_{RBox}} H(p_i, p_i^*) + \frac{1}{|\Omega_{heatmap}|} \sum_{i \in \Omega_{heatmap}} H(p_i, p_i^*) \quad (1)$$

where $|\cdot|$ is the number of positive text pixels in a set, and $H(p_i, p_i^*)$ represents the cross entropy loss between the $i$-th pixel prediction label $p_i$, and its ground truth label $p_i^*$.

We also conduct OHEM to calculate text bounding box regression loss same as FOTS. We select 128 hard positive text pixels and 128 random positive text pixels from each image for regression training. Denoting pixel set selected for text bounding box regression training as $\Omega_{loc}$, the loss function for text bounding box regression can be formulated as:

$$L_{p\_loc} = \frac{1}{N_{pos}} \sum_{i \in \Omega_{loc}} \text{IoU}(R_i, R_i^*) + \lambda_\theta (1 - \cos(\theta_i, \theta_i^*)), \quad (2)$$

where the first term $IoU(R_i, R_i^*)$ is the IoU loss [24] between the predicted text bounding box $R_i$ at the $i$-th pixel and its ground truth $R_i^*$, and the second term is the angle loss [17] between the predicted orientation $\theta_i$ and the ground truth orientation $\theta_i^*$, $\lambda_\theta$ is a weight to balance IoU loss and angle loss, and is set to 10 in our experiments. $N_{pos}$ is the number of positive text pixels.

Therefore, for the pixel-based module, its loss can be formulated as:

$$L_{p\_dt} = L_{p\_cls} + \alpha_p L_{p\_loc}, \quad (3)$$

where $\alpha_p$ is a weight to balance the classification loss and the location loss, set to 1.0 in our experiment.

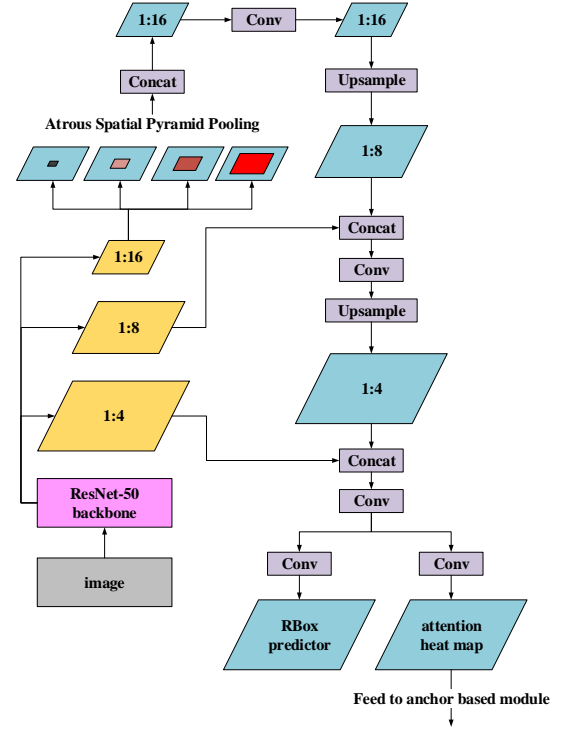

Figure 5: The architecture of the pixel-based module.

## 3.3. The anchor-based module

In the anchor-based module, we modify the SSD

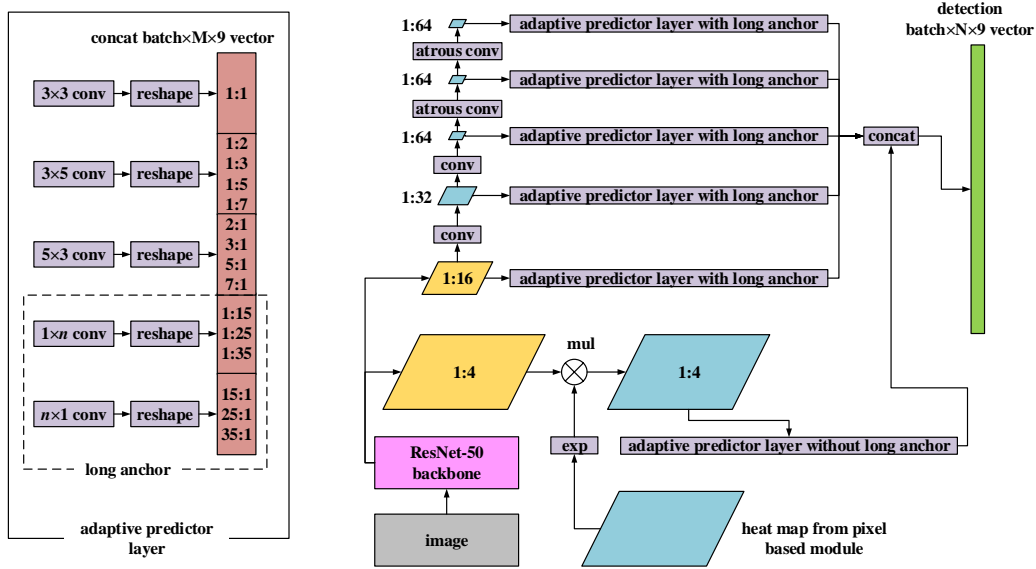

Figure 6: The architecture of anchor based module.

framework to detect scene text having large variances in size and aspect ratio. Our anchor-based module is depicted in Figure 6.

1/4 and 1/16 feature maps from Resnet-50 are used as the backbone, which are shared with the pixel-based module. 1/32, 1/64, 1/64, 1/64 feature maps are appended after 1/16 feature map by fully convolution operation. The six feature maps are denoted as {feat1, feat2, feat3, feat4, feat5, feat6}. To avoid generating too small feature maps, the resolution of the last two feature maps remains unchanged, and the corresponding atrous convolution (rate=2) is applied. The 1/4 feature map instead of the 1/8 one is in the original SSD to enhance the ability of the network to detect small texts. On feat1, attention supervision information from the attention heat map of the pixel-based module is applied. The attention heat map is fed into an exponential operation and then dot with feat1. Using the exponential operation, the probability of each pixel being a positive text is mapped to the range [1.0, $e$], thus able to retain background information and simultaneously highlights the detection information. False positive detections of small texts are reduced in this way.

Further, we propose "Adaptive Predictor Layer" (APL), which is appended to each feature map to get the final text box predictions. In the APL, the anchors are grouped according to their aspect ratios, and different convolutional filters are adopted in each group. Specifically, the anchors are grouped into 5 categories:

a) Square anchors: aspect ratio = 1:1, convolutional filter size 3×3.
b) Medium horizontal anchors: aspect ratios = {1:2, 1:3, 1:5, 1:7}, convolutional filter size 3×5.
c) Medium vertical anchors: aspect ratios = {2:1, 3:1, 5:1, 7:1}, convolutional filter size 5×3.
d) Long horizontal anchors: aspect ratios = {1:15, 1:25, 1:35}, convolutional filter size 1×$n$.
e) Long vertical anchors: aspect ratios = {15:1, 25:1, 35:1}, convolutional filter size $n$×1.

For the long anchors, the parameter $n$ is different on each feature map, and depends on the length of the text line to be detect. For feat1, we trim off the long anchors part in APL, and from feat2 to feat6, $n$ is set to {33, 29, 15, 15, 15} respectively. By using APL, the receptive field of the convolutional filters can better fit the words with different aspect ratios.

Furthermore, to detect dense texts, we propose a definition named "anchor density", as depicted in Figure 7.

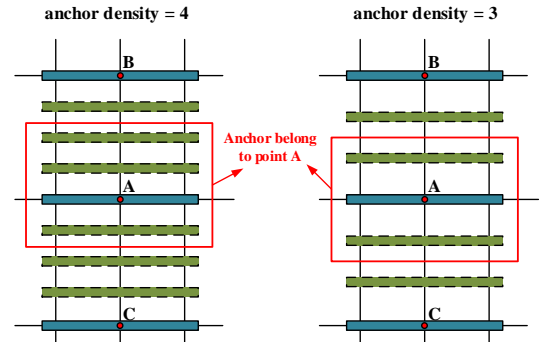

Figure 7: The definition of anchor density.

Each anchor is duplicated with some offsets based on anchor density to better cover dense texts. In Textboxes++, anchors are only duplicated once in vertical direction. In our framework, square anchors are duplicated in both horizontal and vertical directions, the horizontal anchors are duplicated in the vertical direction, and the vertical anchors are duplicated in the horizontal direction. The anchor density is specified for each feature map separately. In our experiments, for the medium anchors, from feat1 to feat6, the anchor densities are {1, 2, 3, 4, 3, 2}, and for the long anchors, from feat2 to feat6, the anchor densities are {4, 4, 6, 4, 3}.

The ground truth label assignment strategy and the loss function of the anchor-based module are similar with those in Textboxes++. The arbitrary-oriented texts are represented by quadrilaterals, and the minimum bounding rectangles (MBRs) of the quadrilaterals are used to match the anchors. Specifically, the anchors are assigned to ground-truth text quadrilaterals (positive text anchors) with the highest IoU larger than 0.5, and to background (negative text anchors) if the highest IoU is less than 0.5.

The output of the anchor-based module is same with TextBoxes++: 9-channel prediction vector for each anchor. The first channel is the probability of each anchor being a positive text quadrilateral, and the following 8 channels predict the expected text quadrilateral's coordinate offsets relative to the anchor.

We adopt OHEM to calculate classification loss, and set the ratio between the negatives and positives to 3:1. Denoting anchor set (negative text anchors and positive text anchors) selected for classification training as $\Omega_a$, the classification loss is formulated as:

$$L_{a\_cls} = \frac{1}{|\Omega_a|}(\sum_{i \in \Omega_a} H(p_i, p_i^*)), \qquad (4)$$

where $|\cdot|$ is the number of positive text anchors in a set, and $H(p_i, p_i^*)$ represents the cross entropy loss between the $i$-th anchor label prediction $p_i$ and its ground truth label $p_i^*$.

The regress loss of the expected text quadrilateral coordinate offsets relative to anchors is formulated as:

$$L_{a\_loc} = \frac{1}{|\Omega_a|}(\sum_{i \in pos(\Omega_a)} SL(l_i, l_i^*)), \qquad (5)$$

where $pos(\Omega_a)$ is the positive part of $\Omega_a$, $SL(l_i, l_i^*)$ represents the smooth L1 loss [25] between the predicted offset coordinate $l_i$ and the ground truth $l_i^*$ for $i$-th anchor.

Therefore, for the anchor-based module, its loss can be formulated as:

$$L_{a\_dt} = L_{a\_cls} + \alpha_a L_{a\_loc}, \qquad (6)$$

where $\alpha_a$ is a weight to balance the classification loss and the location loss, and is set to 0.2 for quick convergence in our experiment.

### 3.4. Training and inference

At the stage of training, the whole network is trained end-to-end using ADAM [26] optimizer, and the loss of the model can be formulated as:

$$L_{all} = (\alpha_{all}L_{p\_dt} + L_{a\_dt}), \qquad (7)$$

where $\alpha_{all}$ is a weight to balance the pixel-based loss and the anchor-based loss, set to 3.0 in our experiments. For data augmentation, we uniformly sample 640×640 crops from images to form a mini-batch of size 32. The model trained on ImageNet dataset [27] is adopted as our pre-trained model. The SynthText dataset [28] containing 800k synthesized text images is used for pre-training our model, then the training process is continued on the corresponding images of each benchmark dataset. For each dataset, the initial learning rate is set to 0.0001 at the first stage training, then reduced to 0.00001 at the second stage training.

At the stage of inference, we propose the "fusion NMS" to get the final detection results. We assign the anchor-based module to detect small texts and long texts and assign the pixel-based module to detect medium size texts. In the APL of the anchor-based module, anchors trimming is conducted. All the anchors on the 1/4 feature map (i.e., Category a, b, and c mentioned in Section 3.3) and all the long anchors on the other feature maps (i.e., Category d and e mentioned in Section 3.3) are retained. Because the anchors on the 1/4 feature map (generally small in size) usually have no enough room to contain two large angle text instances, and the long anchors can only match with small angle text instances, "Anchor Matching Dilemma" is much less likely to happen. In the pixel-based module, we filter out the predicted text boxes when the minimum size of its MBR is less than 10 pixels and the aspect ratio of its MBR is not in the range [1:15, 15:1]. Finally, we gather all the remaining candidate text boxes, and conduct a cascaded NMS similar to Textboxes++ to get the final detection results. The NMS is first applied with a relatively high IOU threshold (e.g. 0.5) on the MBRs of the predicted text quadrilaterals. This operation on the MBRs is much less time-consuming and removes most of the candidate boxes. Then the time-consuming NMS on text quadrilaterals is applied to the remaining candidate boxes with a lower IOU threshold (e.g. 0.2). Because the candidate text boxes from the anchor-based module and the pixel-based module are overlapped, we add 1.0 to the scores of text boxes predicted by the anchor-based module, those text boxes have higher priority when conducting the NMS.

### 4. Experiments

We first use the public SynthText dataset and our own poster dataset for the characteristics of the model, then evaluate our method on two challenging public benchmarks: ICDAR 2015 [29] and ICDAR 2017 MLT [30].

### 4.1. The ability of detecting small scene text

To demonstrate that the performance of our method is better than the pixel-based method in detecting small texts, we conduct experiments on SynthText dataset. In both training and inference, the resolution of images is resized to 384×384 with reserved height-width ratio and padded short side. We randomly select 4000 images as the validation dataset, and compare Pixel-Anchor, the pixel-based method, and the anchor-based method in Table 1. Pixel-Anchor outperforms the pixel-based method and the anchor-based method in F-measure. The anchor-based method has high recall scores because it performs better in detecting small texts. Figure 8 shows the detection details of the two methods. Taking advantage of the small anchors, Pixel-Anchor can predict the locations more accurately for small texts. The anchor-based method itself is less accurate than the pixel-based method, but by combining the anchor-based module and the pixel-based module, Pixel-Anchor can maintain both high precision score and high recall score.

Table 1: Results of comparison in detecting small images.

| Method | Resolution | Precision | Recall | F-measure |
|---|---|---|---|---|
| Pixel-Anchor | | 0.962 | 0.902 | 0.931 |
| Pixel only | 384×384 | 0.962 | 0.872 | 0.915 |
| Anchor only | | 0.943 | 0.898 | 0.920 |

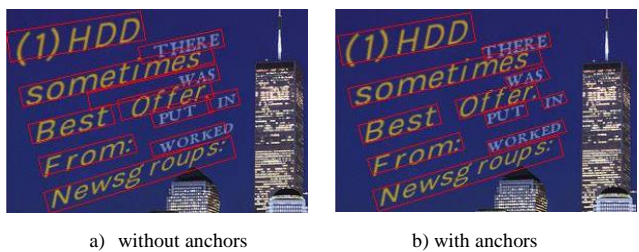

a) without anchors     b) with anchors

Figure 8: With and without the anchor-based module in detecting small texts (we crop only a small part of the original images for illustration).

4.2. The ability of detecting dense large-angle texts

We compare Pixel-Anchor and the anchor-based method in detecting dense large-angle texts, shown as Figure 9. The anchor-based method alone performs badly in detecting dense blocks of large-angle texts due to the anchor matching dilemma.

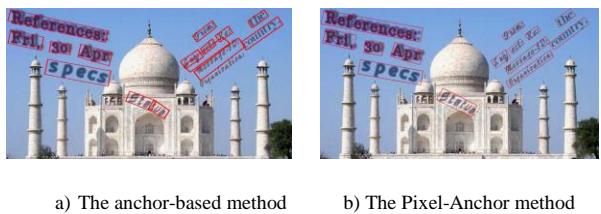

a) The anchor-based method     b) The Pixel-Anchor method

Figure 9: Comparison with and without pixel-based module on detecting dense large-angle texts.

4.3. The ability of detecting long text lines

To demonstrate that our proposed APL can effectively detect long text lines, we test Pixel-Anchor using our poster dataset, which contains numerous long Chinese text lines running across images. We collect and label 5000 poster images, select 4000 images as training dataset, and 1000 images as validation dataset. On the poster dataset, we achieve an F-measure of 0.88 for 768×768 resolution image. Shown as Figure 10, Pixel-Anchor performs well in detecting dense and long Chinese text lines. Compared with the methods like Pixel-Link and CTPN [31], using APL to detect long text lines is outstanding. In the mentioned methods, the linkages between pixels (or segments) are predicted after which extensive post-processing are applied to join pixels or segments together. In our method, we only conduct a simple NMS in post-processing, which is more robust than the linking methods. Compared with the pixel-based method, Pixel-Anchor prevails in detecting small, dense, and long text lines, and the pixel-based method tends to recognize two close text lines as one.

4.4. Evaluation on the public datasets

Pixel-Anchor is evaluated on two challenging benchmark datasets, ICDAR 2015 and ICDAR 2017 MLT, and outperforms the competing methods in terms of text localization accuracy and run speed. Some detection results are shown in Figure 11. For fair comparisons, only the single-scale results on the two datasets are reported.

**ICDAR 2015** is the challenge 4 of ICDAR 2015 Robust Reading Competition, which is commonly used for oriented scene text detection. This dataset includes 1000 training images and 500 testing images. Because these images are captured by google glasses without taking care of position, the texts can be in arbitrary orientations. Some blurred text regions in ICDAR 2015 datasets are labeled as "DO NOT CARE", and we ignore them in training.

As shown in Table 2, our approach on ICDAR 2015 gets a competitive result, the proposed algorithm achieves an F-score of 0.8768 at 10 FPS for a 960×1728 resolution image on an Nvidia Titan X GPU. The FOTS with recognition branch performs slightly better than our method. However, it uses text recognition supervision to help the network to perform better in detection. If only the single-scale results without the help of recognition are taken into account, to the best of our knowledge, our result 0.8768 is the best reported result in literature. Compared with other methods, our networks achieve a higher recall score.

**ICDAR 2017 MLT** is a large scale multi-lingual text dataset, which includes 7200 training images, 1800 validation images, and 9000 testing images. The dataset is composed of challenging scene images which include 9 languages and text regions in this dataset can be in arbitrary orientations, leading to higher diversity and difficulty.

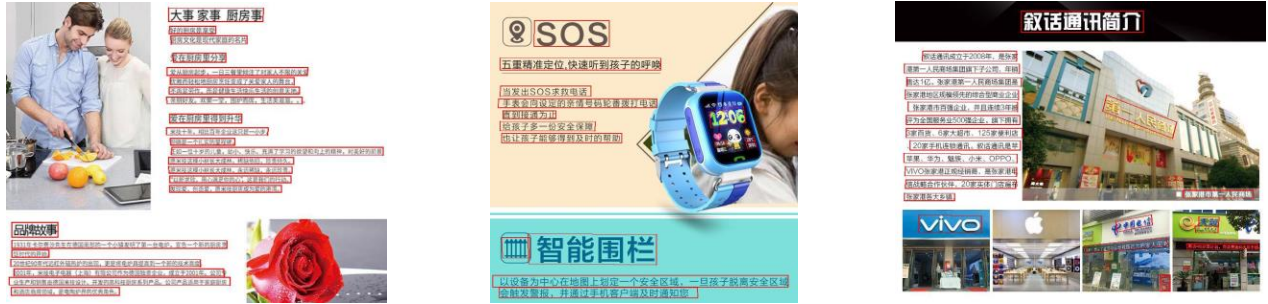

Figure 10: Some examples of detection results on our poster dataset.

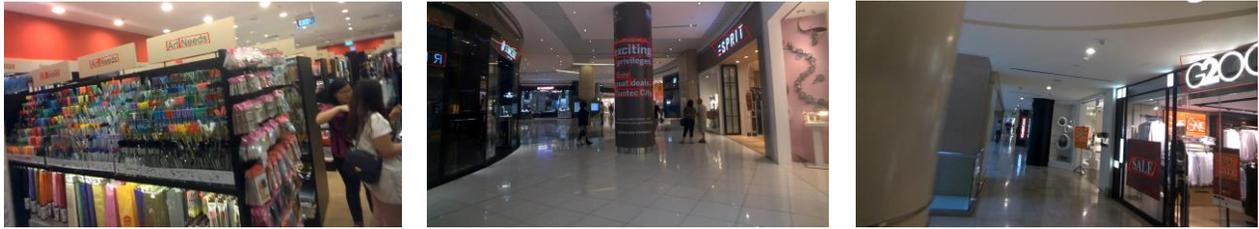

Figure 11: Qualitative results of the proposed algorithm on ICDAR 2015.

As shown in Table 3, our method on ICDAR 2017 MLT achieves an F-score of 0.681 for an 1152×1152 resolution image. In table 3, PSENet performs better than our method due to a different mechanism. The model resizes the longer sides of input images to 3200 pixels and adopts a pixel-linking post-processing without limit of receptive field. For the multi-scale results, our method also ranks first in literature. The outcome is: precision: 83.90; recall: 65.80; F-score: 73.76.

Table 2: The single-scale results on ICDAR 2015.

| Method | Precision | Recall | F-score |
| --- | --- | --- | --- |
| CTPN [31] | 74.22 | 51.56 | 60.85 |
| SegLink [32] | 74.74 | 76.50 | 75.61 |
| SSTD [33] | 80.23 | 73.86 | 76.91 |
| WordSup [34] | 79.33 | 77.03 | 78.16 |
| EAST [17] | 83.27 | 78.33 | 80.72 |
| TextBoxes++ [14] | 87.2 | 76.7 | 81.7 |
| R$^2$CNN [35] | 85.62 | 79.68 | 82.54 |
| Pixel-Link [18] | 85.5 | 82.0 | 83.7 |
| FTSN [36] | 88.65 | 80.07 | 84.14 |
| SLPR [37] | 85.5 | 83.6 | 84.5 |
| Incept Text [38] | 90.5 | 80.6 | 85.3 |
| FOTS [20] detection branch only | 88.84 | 82.04 | 85.31 |
| Mask Text [39] | 91.6 | 81.0 | 86.0 |
| PSENet [19] | 89.30 | 85.22 | 87.21 |
| FOTS [20] with recognition branch | 91.0 | 85.17 | 87.99 |
| Pixel-Anchor | 88.32 | 87.05 | 87.68 |

Table 3: The single-scale results on ICDAR 2017 MLT.

| Method | Precision | Recall | F-score |
| --- | --- | --- | --- |
| SARI_FDU_RRPN_v1 [41] | 71.17 | 55.50 | 62.37 |
| Lyu et al. [42] | 83.8 | 55.6 | 66.8 |
| FOTS [20] detection branch only | 79.48 | 57.45 | 66.69 |
| FOTS [20] with recognition branch | 80.95 | 57.51 | 67.25 |
| PSENet [19] | 77.01 | 68.40 | 72.45 |
| Pixel-Anchor | 79.54 | 59.54 | 68.10 |

5. Conclusion

In this paper, we propose a novel end-to-end trainable deep neural network framework named Pixel-Anchor, which combines semantic segmentation and SSD in one network by feature sharing and the anchor-level attention mechanism for oriented scene text detection. Our method combines the advantages of the pixel-based method and the anchor-based method and avoids their shortcomings. In the anchor-based module of Pixel-Anchor, we propose APL for SSD to better detect objects which have a variety of aspect ratios like scene text. With APL, dense blocks of small text and long text lines can be detected more efficiently. The impressive performances on the public datasets demonstrate the effectiveness and robustness of our method for scene text detection.